\documentclass[10pt, letterpaper, twocolumn, conference]{IEEEtran}
\usepackage{graphicx,times,amsmath}
\usepackage{cases}
\usepackage{multirow}
\usepackage{algorithm}
\usepackage{algorithmic}
\usepackage{amsfonts}
\usepackage{graphicx}
\usepackage{amssymb}
\usepackage{amsfonts}
\usepackage{subfig}
\usepackage{epstopdf}
\usepackage{makecell}
\usepackage{array}
\usepackage{xspace}
\usepackage{xparse}
\usepackage{mathtools}
\usepackage{cite}
\usepackage[symbol]{footmisc}

\newcolumntype{?}{!{\vrule width 1pt}}

\DeclareMathOperator*{\argmin}{arg\,min}
\DeclarePairedDelimiter\ceil{\lceil}{\rceil}

\usepackage[colorlinks=true,citecolor={black},breaklinks=true,letterpaper=true,bookmarks=false]{hyperref}

\newcommand{\XL}[1]{{#1}}
\makeatletter
\DeclareRobustCommand\onedot{\futurelet\@let@token\@onedot}
\def\@onedot{\ifx\@let@token.\else.\null\fi\xspace}
\def\eg{\emph{e.g}\onedot} 
\def\ie{\emph{i.e}\onedot} 
 
\def\etc{\emph{etc}\onedot}

\makeatother

\begin{document}
\setlength{\abovedisplayskip}{0.3pt}
\setlength{\belowdisplayskip}{0.3pt}
\title{Hybrid Camera Pose Estimation with Online Partitioning for SLAM \vspace{-5pt}
}

\DeclareRobustCommand*{\IEEEauthorrefmark}[1]{%
  \raisebox{0pt}[0pt][0pt]{\textsuperscript{\footnotesize #1}}%
}
\author{
{Xinyi Li\IEEEauthorrefmark{1}
\thanks{This work was supported in part by US NSF (Grants 1814745, 1618398 and 2002434), the Yahoo Faculty Research and Engagement Program Award, and the Amazon AWS Machine Learning Research Award.}
\thanks{{\IEEEauthorrefmark{1}}Xinyi Li is with Department of Computer and Information Sciences, Temple University, Philadelphia, USA, \{{\tt\small xinyi.Li@temple.edu}\}} 
and Haibin Ling\IEEEauthorrefmark{2}\thanks{{\IEEEauthorrefmark{2}}Haibin Ling is with Department of Computer Science, Stony Brook University, Stony Brook, USA, \{{\tt\small hling@cs.stonybrook.edu}\}}}\vspace{-5pt}
}
\maketitle

\begin{abstract}
     This paper presents a hybrid real-time camera pose estimation framework with a novel partitioning scheme and introduces motion averaging to monocular Simultaneous Localization and Mapping (SLAM) systems. Breaking through the limitations of fixed-size temporal partitioning in many conventional SLAM pipelines, our approach significantly improves the accuracy of local bundle adjustment by gathering spatially-strongly-connected cameras into each block. With the dynamic initialization using intermediate computation values, \XL{we improve the Levenberg-Marquardt solver to further enhance the efficiency of the local optimization.} Moreover, the dense data association between blocks by  our co-visibility-based partitioning enables us to explore and implement motion averaging to efficiently align the blocks globally, updating camera motion estimations on-the-fly. Experiments on benchmarks convincingly demonstrate the practicality and robustness of our proposed approach by significantly outperforming conventional approaches.  
\end{abstract}

\section{Overview}
\subsection{Introduction}
With low-cost hand-held scanning devices made widely available recently, simultaneous localization and mapping (SLAM) and structure-from-motion (SfM) systems have been extensively studied in the past decades. Camera pose estimation, lying at the core of most feature-based multi-view geometry systems, aims to jointly refine the coordinates of 3D landmarks and camera parameters along the camera trajectory. 

This paper proposes a novel hybrid formulation to address the real-time camera pose estimation problem in SLAM. To the best of our knowledge, our approach is the first hybrid SLAM framework combining {\it bundle adjustment} (BA) and {\it motion averaging} to estimate and update the camera poses on-the-fly. As shown in the experiments, our approach clearly outperforms state-of-the-art conventional BA-based systems in both efficiency and accuracy. Experimental results on large-scale outdoor scene dataset further demonstrates the robustness of our proposed framework.

\begin{figure*}[th!]
\centering
\includegraphics[width=\textwidth]{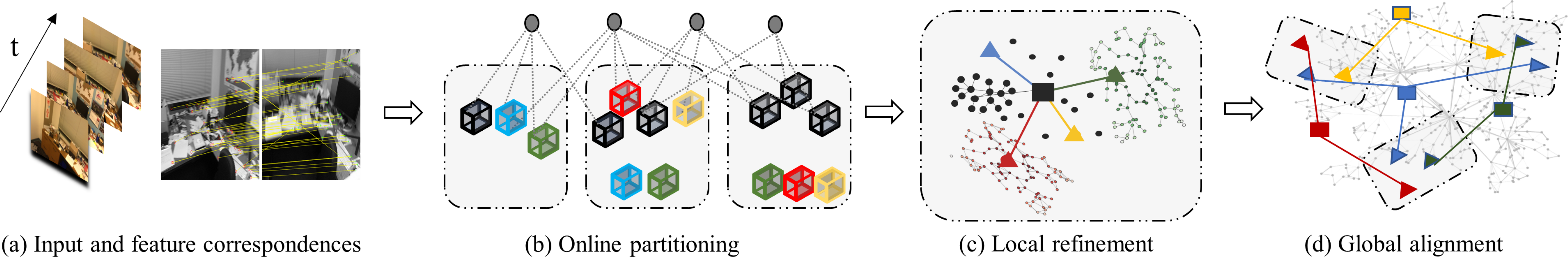}
\caption{\small The pipeline of our proposed algorithm: (a) the input image sequence and feature correspondences, (b) the online partitioning based on local and global co-visibility (\S\ref{sec:partition}), (c) the refinement of local pose-graph with semi-distributed initialization (\S\ref{sec:local}), and (d) the progressive global alignment by motion averaging with common camera poses (\S\ref{sec:global}).}
\label{fig:pipeline}
\vspace{-10pt}
\end{figure*}

\subsection{Related Work}
Comprehensive existing approaches have been proposed to address the accuracy and efficiency of SLAM. Considered as the arguably most popular approach, BA has received extensive research attention. For instance, {\it incremental BA} (\cite{Indelman2012a, Vo2016b, Sibley2009b, Wu2013b,schonberger2016structure, Liu2018a}) starts by initializing the map reconstruction with few images and later measurements are incrementally merged with iterative essential matrix updates. Many state-of-the-art SLAM systems (\cite{Konolige2008b,kaess2008,ila2017}) implement incremental BA to compromise speed and accuracy. However, incremental approaches are well-observed to be prone to accumulated drifting errors and sensitive to poor initialization, resulting in the deteriorated performance progressively over long camera trajectories. Alternatively, most approaches based on
{\it hierarchical BA} (\cite{Shum1999,Maier2014d,Ni2007a, steedly2003spectral, havlena2010efficient,cui2017hsfm}) pre-cluster the images and then solve smaller sub-problems separately. Though achieving excellent results for the off-line SfM problem, hierarchical approaches require global view-graph information in advance such that the applicability for generalized real-time systems like SLAM is hindered. 
Many other challenging issues of BA remain open, \eg, the cubic-order complexity with regards to the number of image frames involves prohibitive computational costs~\cite{Wu2013b}; slow or erroneous convergences of the numerical solvers due to the intrinsically high dimensional solution space, \etc. Both~\cite{grisetti2012robust} and~\cite{ni2007tectonic} implement divide-and-conquer schemes in SLAM, addressing the aforementioned issues: \cite{grisetti2012robust} offers a robust approach but lacks accuracy analysis, while~\cite{ni2007tectonic} requires manual partitioning that is not always practical. To tackle these problems, this paper presents a hybrid pose estimation framework combining BA and motion averaging for feature-based monocular SLAM.

With the efficient implementations of motion averaging (\cite{cui2015,zhu2018,zhu2017parallel, chatterjee2013efficient, chatterjee2018robust, hartley2013rotation, fredriksson2012simultaneous}), promising results have recently been shown in off-line distributed SfM frameworks (\cite{cui2015, cui2017csfm, eriksson2016consensus, zhang2017distributed, zhu2017parallel, zhu2018}). \cite{zhu2017parallel} proposes a camera clustering algorithm and presents a hybrid pipeline applying the parallel-processed local increment into global motion averaging framework. In~\cite{zhu2018}, a divide-and-conquer framework is introduced to address distributed large-scale motion averaging problems and shows convincing results on city-scale datasets. Inspired by the superior performances, we propose a {\it quasi-distributed} formulation and introduce {\it single rotation averaging} into real-time pose estimation tasks.
 
Previous works most related to ours include~\cite{cui2017hsfm,locherprogsfm,alvarowhyba}, among which \cite{cui2017hsfm} and \cite{locherprogsfm} both aim to solve the off-line SfM problem. In~\cite{cui2017hsfm}, a unified framework is proposed where global rotation averaging is implemented in a community-based structure, followed by local incremental BA for individual camera refinement. Relaxing the requirement in~\cite{cui2017hsfm} of prior global knowledge for the complete view graph, \cite{locherprogsfm} is capable of processing both unordered and sequential data. However, it is impractical to directly apply~\cite{locherprogsfm} to on-line systems since 1) the progressive brute-force image matching for clustering struggles to deliver real-time performance; and 2) the global multiple similarity averaging based on dense inter-cluster connections reaches the memory and computation limits of real-time systems very soon.~\cite{alvarowhyba} presents a motivating BA-free SLAM framework where only camera orientations are estimated and maintained via rotation averaging. However, global optimality of the camera pose parameters in the proposed framework is not claimed and the accuracy compared with state-of-the-art systems is not provided. \subsection{Our Approach}
In order to break through the limitations of fixed-size partitioning in the conventional framework, we instead partition cameras into blocks based on both temporal and spatial co-visibility. Specifically, in addition to temporally-adjacent cameras, each block contains cameras added from previous blocks if sufficient overlapping views are detected between the cameras and the current block. 

The local optimization process in conventional BA refines the camera poses within sequential partitions of fixed sizes, and frequently suffers from slow or erroneous convergences due to noisy data. By contrast, our partitioning scheme gathers spatially-strongly-connected intra-block cameras to improve the accuracy of the local optimization afterwards. It also encourages the propagation of intermediate computation values between blocks, and later benefits the local BA in two-fold: 1) good initialization is dynamically provided; and 2) duplicate computations are effectively avoided. The satisfying initialization also sheds light on our proposed self-adaptive Levenberg-Marquardt (L-M) solver, which achieves a quadratic convergence rate and thus further enhances the local BA efficiency.  

Shared camera poses among different blocks, produced by our partitioning scheme, allow us to close the gaps between off-line and on-line systems and introduce motion averaging into solving real-time pose estimation tasks. By leveraging single rotation averaging on the common camera poses in multiple blocks, we effectively reduce drifting errors while tackling the computational bottleneck of conventional formulation of global BA. Moreover, by virtue of the dense connectivity between blocks based on global co-visibility, our approach can skip explicit loop closing to further accelerate the global alignment.

In summary, the key contributions of our paper are:
\begin{itemize}
\item We develop a co-visibility-based partitioning scheme according to both temporal and spatial feature correspondences, as to effectively improve the accuracy of local optimization while guaranteeing the global consistency.

\item We introduce motion averaging to efficiently update the camera poses on-the-fly, conquering the computational bottleneck of conventional global BA and avoiding explicit loop closing.
\item With the dynamic inter-block propagation and intra-block broadcasting of intermediate computation values, we introduce \XL{a novel update law for the damping parameter in the L-M solver, designed for approaches based on varying sized sub-maps to further benefit the convergence.} 
\end{itemize}


\section{Preliminaries and Notations}
\label{sec:overview}
\textit{Bundle adjustment} (BA) is a fundamental technique in SLAM and SfM. By minimizing the sum of re-projection errors for given observations of the 3D scene, BA iteratively refines camera parameters and 3D point coordinates jointly. 

Consider a given 3D scene with $m$ frames and $n$ observable 3D points. Let $p_i^j\in \mathbb{R}^2$ denote the observation coordinate of 3D point $P_j\in \mathbb{R}^3$ on image frame $I_i$ for $1\leq i\leq m$, $1\leq j\leq n$. We define the projection mapping $\pi (K,C_i, P_j):\mathcal{K}\times \mathcal{C} \times \mathbb{R}^3 \mapsto \mathbb{R}^2$, where $K\in \mathcal{K}\subseteq \mathbb{R}^{3\time 3}$ denotes the camera intrinsic matrix and is considered known and fixed for a calibrated system; $C_i =[R_i|t_i]\in \mathcal{C}\subseteq \mathbb{SE}(3)\subseteq \mathbb{R}^{3\times 4}$ denotes the absolute camera pose of $I_i$, where the rotation is denoted by $R_i\in \mathbb{SO}(3)\subseteq \mathbb{R}^{3\times 3}$ and  translation by $t_i\in \mathbb{R}^3$. The re-projection error minimization is then
\begin{equation}
    \label{eq:repro}
    \min_{\substack {C_i \in \mathcal{C}, 1\leq i\leq m \\ P_j, 1\leq j\leq n}} \sum_{i=1}^{m} \sum_{j=1}^n \mathbb{I}_{ij} \|p_i^j - \pi(K, C_i, P_j)\|_2^2,
\end{equation}
where $\mathbb{I}$ denotes the visibility matrix, \ie, $\mathbb{I}_{ij}=1$ if $P_j$ is visible on $I_i$ and $0$ otherwise. In \S\ref{sec:local} we describe the local optimization process by solving sub-problems of Eq.~\ref{eq:repro}. 

\textit{Motion averaging} aims to optimize the camera motion based on several known camera orientations. We leverage the {\it single rotation averaging} to align the camera blocks globally, where single rotations are computed given multiple measurements by averaging. 
Formally, let $R_{ii'}$ denote the relative rotation from $I_i$ to $I_{i'}$, \ie $R_{ii'}=R_{i'} R_i ^\top$, we define the optimization as
\begin{equation}
\label{eq:rotavg}
    \argmin _{R_{ii'}\in \mathbb{SO}(3)}\sum_{k\in \mathbb{N}}d\Big(\widetilde{R_{ii'}^k} , R_{ii'}\Big)^2,
\end{equation}
where $\widetilde{R_{ii'}^k}\in \mathbb{SO}(3)$ denotes the estimation of $R_{ii'}$ by the $k^\text{th}$ observation; $d(\cdot , \cdot)$ denotes the proper metric, in this paper we use the Karcher mean~\cite{grove1973conjugatec}, \ie, the geodesic $L_2$- mean for the computation. Formally, define the metric $d(X,Y)=d_{\angle}\big(XY^\top , I\big)=\|\log \big(XY^\top\big)\|_2$, $\forall X,Y\in \mathbb{SO}(3)$. Further implementation details are illustrated in \S\ref{sec:global}.
\section{The Hybrid Formulation}
\label{sec:saba}
In this section, we propose to dynamically partition the cameras into blocks according to both temporal and spatial co-visibility, producing blocks of spatially-strongly-connected cameras (\S\ref{sec:partition}). Followed by the local optimization (\S\ref{sec:local}), we then propagate the intermediate computation values of the common camera poses. Single rotation averaging is performed afterwards on the shared camera poses among blocks in the global alignment (\S\ref{sec:global}). For the sake of simplicity, we refer camera(s) to the absolute camera pose(s) throughout the paper.
\subsection{Online Partitioning}
\label{sec:partition}
Key to our proposed algorithm is to timely partition the incoming sequence into blocks where cameras within the same block are spatially-strongly-connected regarding co-visible 3D scene points. Precisely, we enforce that most 3D points are visible by a sufficient amount of cameras to ensure the intra-block camera-camera connectivity. Besides, we allow cameras from previous sequential blocks to be added to current block if sufficient camera-block overlapping views are detected.

 Given the input sequence of $m$ image frames, each frame $I_i$ is parameterized by $\{C_i, \mathbf{P}_i\}$ with $\mathbf{P}_i= \{P_j|\mathbb{I}_{ij}\neq 0\}$ as the set of visible 3D points on $I_i$. We aim to partition the sequence into $m_b$ blocks $\mathcal{B}^l=\{\mathcal{C}^l,\mathcal{P}^l\},1\leq l\leq m_b$, assuming $m^l$ cameras are partitioned into $\mathcal{B}^l$, then $\mathcal{C}^l=\{C_i^l\in \mathcal{C}\}_{i=1}^{m_l}$ denotes the set of cameras and $\mathcal{P}^l=\cup_{i=1}^{m_l} \mathbf{P}_i $ denotes the union set of 3D points visible by the entire $\mathcal{C}^l$. Since we encourage cameras to appear in multiple blocks, the blocks produced by our partitioning scheme satisfy the following
 \begin{equation}
 \label{eq:partitionset}
 \begin{split}
 \small
      \cup_{l=1}^{m_b} \mathcal{C}^l = \cup_{i=1}^m C_i, \hspace{2pt} \cup_{l=1}^{m_b} \mathcal{P}^l = \cup_{i=1}^m\mathbf{P}_i,\\
      \exists 1\leq l' \neq l \leq m_b \text{  s.t.  } \mathcal{C}^l \cap \mathcal{C}^{l'} \neq \emptyset,
 \end{split}
 \end{equation}
 \ie, for each block, there exists at least one block such that the two blocks have common cameras. In details, our proposed partitioning criterion consists of two parts as described in the following discussions, namely, the \textit{local co-visibility} and the \textit{global co-visibility}.
 
The local co-visibility is defined as the average degree of connectivity between the cameras within a block. Specifically, the accuracy of the intra-block BA can be asserted with each single 3D point visible by a sufficient number of cameras. However, it is implausible to employ this principle in real-time applications since it can introduce a large number of redundant cameras. Alternatively, we define the \textit{local co-visibility score} $\gamma^l$ for $\mathcal{B}^l$ by averaging the camera numbers over the 3D point set, and use it to guide the block construction. Formally, let $\mathbf{p}_j^l$ denote the union set of projected 2D points on all in-block camera frames from a 3D point $P_j \in \mathcal{P}^l$, then
\begin{equation}
    \label{eq:gamma}
    \gamma ^l = {\textstyle\sum_{j=1}^{|\mathcal{P}^l|}\big|\mathbf{p}_j^l\big|} \Big/ {|\mathcal{P}^l|}, 
\end{equation}
where $|\cdot|$ denotes the size of a set. We require that $\gamma ^l \geq \gamma^\text{thr}$, where $\gamma^\text{thr}\geq 3$ is a preset parameter determining the average size of the blocks, \eg, with a higher value of $\gamma^\text{thr}$, blocks with greater sizes are expected. By enforcing a minimum local average co-visibility, the accuracy of local BA inside each block is improved with sufficiently many point-camera constraints. New image frames are sequentially partitioned into current block $\mathcal{B}^l$ while $\gamma^l<\gamma^\text{thr}$, we therefore limit the size of each block by $n^\text{thr}$ in order to keep each block small for the real-time performance. Note that for scenarios where the camera is moving fast in feature-less scenes, our approach degrades to the conventional fixed-size partitioning. Once the local co-visibility score meets the criterion, block is temporally fixed followed by the add-in's of the image frames satisfying the global co-visibility requirement as illustrated below.
\begin{figure}[!t]
    \centering
    \includegraphics[width=\linewidth]{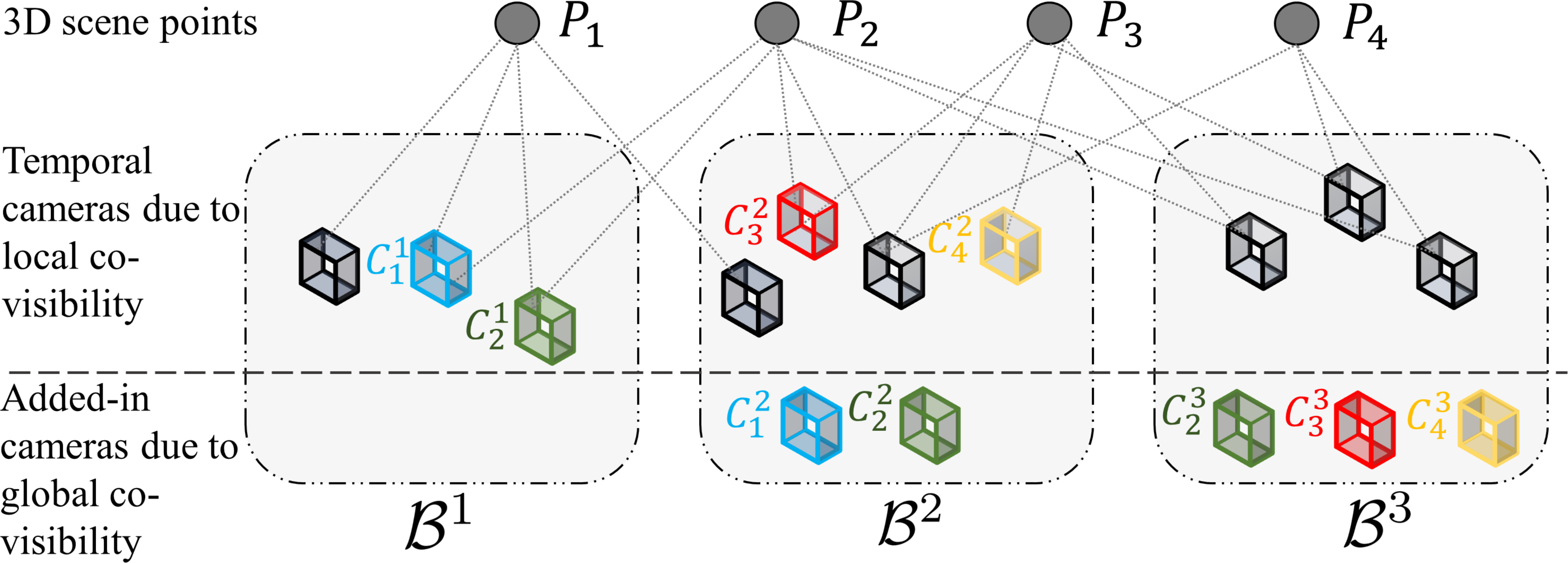}
    \caption{\small Our online partitioning scheme partitions the cameras into blocks based on both temporal and spatial co-visibility. The key is that one camera pose can be optimized within different blocks (\eg $C_1$ is estimated in $\mathcal{B}^1$ and $\mathcal{B}^2$, marked as $C_1^1$ and $C_1^2$, respectively) as long as it shares sufficient overlapping views with the blocks.}
    \label{fig:partition}
    \vspace{-0.27cm}
\end{figure}

The global co-visibility is defined as the degree of camera-block connectivity, \ie, instead of only optimizing the frames in the current temporal block, cameras with sufficient overlapping views from previous blocks are added into the current block. The {\it global co-visibility score} $\beta_i^l$ for the covisibility between the current block $\mathcal{B}^l$ and a previous image frame $I_i$ (\ie, $C_i \in \mathcal{C}^{l'}, l'<l$) is defined as
\begin{equation}
    \label{eq:beta}
    \beta_i^l = {|\mathbf{P}_i\cap \mathcal{P}^l|} \big/ {|\mathcal{P}^l|} .
\end{equation}

$I_i$ can thus be added into $\mathcal{B}^l$ if $\beta_i^l > \beta^\text{thr}$, where $\beta^\text{thr} \in [0,1)$ is a preset parameter defining the minimum camera-block overlapping view ratio. Furthermore, we set an {\it upper bound} $n_\alpha$ for the added-in cameras, which is particularly important when the capturing density is considerably high (\eg camera moving around small-scale scenes for a long duration), as the high volume of overlapping visible regions can introduce an excessive amount of cameras therefore causing noisy data and additional computations.  In practice, higher values of $\gamma ^\text{thr}$, $n_\alpha$ and $n^\text{thr}$ combined with a lower global co-visibility threshold $\beta^\text{thr}$ achieve a better accuracy with trade-offs of a slower convergence and higher computational cost. In our experiments we choose $\gamma ^\text{thr} =10$, $\beta^\text{thr}=0.15$, $n_\alpha=10$ and $n^\text{thr}=50$ to balance between accuracy and efficiency. We illustrate our partitioning scheme in Fig.~\ref{fig:partition}.

Note that two consecutive blocks share {\it at least one} overlapping frame: the last temporal frame in the previous block is the first temporal frame in the current block by default. The current block is considered fixed at the completion of adding cameras from previous blocks, hence proceeding to the following local optimization. Note that the blocks stay unchanged during the optimization, despite that the potential updates on points and camera poses might slightly perturb the co-visibility, to counterbalance the real-time processing speed.

\subsection{Local Optimization}
\label{sec:local}

We first reformulate Eq.~\ref{eq:repro} into the sub-problem for $\mathcal{B}^l$, the objective function of the local optimization is defined as
\begin{equation}
    \label{eq:locrepro}
    \hspace{-2.3mm}(\mathcal{C}^l,\mathcal{P}^l)^* \hspace{-1.mm}=\hspace{-5mm}\argmin_{C_i\in\mathcal{C}^{m_l} ,P_j\in \mathbb{R}^{3\times n_l}} \hspace{-.52mm}\sum_{\substack{C_i \in \mathcal{C}^l\\P_j \in \mathcal{P}^l}}\hspace{-.2mm} \mathbb{I}_{ij} \|p_i^j - \pi(K, C_i, P_j)\|_2^2.
\end{equation}

To iteratively solve Eq.~\ref{eq:locrepro} we follow the Levenberg-Marquardt algorithm~\cite{Levenberg1944}, which is known to be sensitive to initialization. Indeed, conventional BA often suffers from converging to the local minima when a poor starting point is chosen. With our partitioning scheme, we tackle the issue by propagating the intermediate computation values of the parameters of added-in cameras from previous blocks. 

In details, the added-in cameras carry computation values from the previous intra-block local BA, we can thus initialize the local BA iterations of the current block with the intermediate results. Hence the local optimization of each block quasi-distributedly launches as soon as the block is fixed, with no delay on requesting final computation results of previous block. Additionally, we broadcast the intermediate values based on the intra-block camera connectivity greedily. 
\begin{figure}[t!]
\begin{center}
\hspace{-0.5cm}
\subfloat [Average Trajectory Error]{
\includegraphics[width=.497\linewidth]{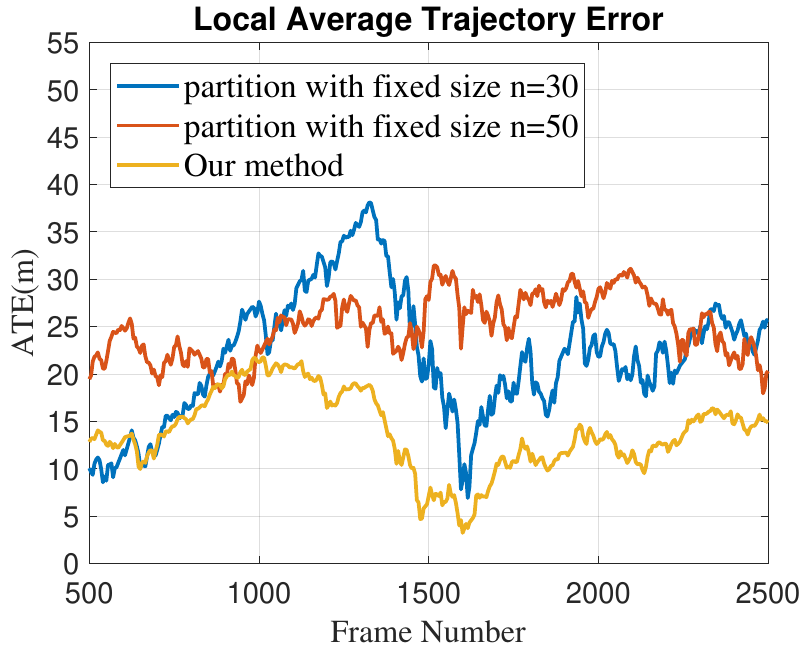}
}
\subfloat [Local Iteration Numbers]{
\includegraphics[width=.488\linewidth]{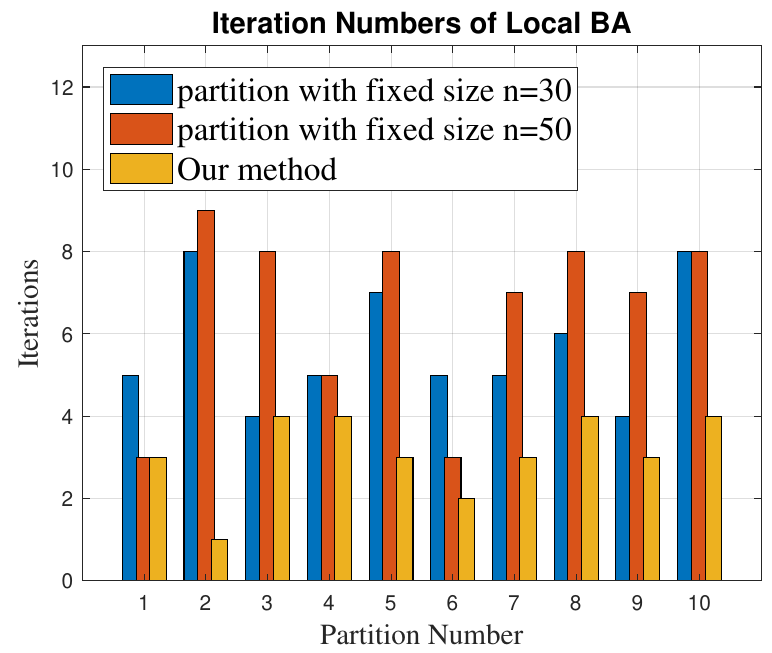}
}
\caption{\small Comparison with conventional local optimizations of fixed-size partitions: (a) the average trajectory error for the local block with the peaks representing that a new block, and (b)  iterations required for the convergence of the intra-block local BA. The experiments are conducted on Seq.02 of KITTI Odometry~\cite{Geiger2013c}. More details can be found in the supplementary materials.}
\label{fig:localcomp}
\end{center}
\vspace{-0.5cm}
\end{figure}

\footnotetext{Supplemental materials available at \protect\url{https://sites.temple.edu/xyli/hybrid/} }
Formally, let $\mathcal{G}^l=\{\mathcal{V}^l, \mathcal{E}^l\}$ denote the intra-block pose-graph for $\mathcal{B}^l$, where the vertices in $\mathcal{V}^l$ represent the cameras and the edges in $\mathcal{E}^l$ are defined by the co-visibility between two image frames, \ie,
\begin{subequations}
\begin{align}
\label{eq:graph}
\mathcal{V}^l&=\{C_i|C_i\in \mathcal{C}^l\},\\
\mathcal{E}^l&=\{e_{ij}|1\leq i\leq m_l,1\leq j\leq n_l\},
\end{align}
\end{subequations}
where $e_{ij}=1$ if there exists co-visible 3D points between frames $I_i$ and $I_j$ and $0$ otherwise. The broadcasting of the parameters is conducted by generating a Maximum Spanning Forest (MSF) of multiple Maximum Spanning Trees (MST)~\cite{kruskal1956shortest} from $\mathcal{G}^l$ as discussed below. 

Let $\mathcal{C}_\alpha ^l$ denote the set of cameras added into $\mathcal{B}^l$ from previous blocks, where $\mathcal{C}_\alpha ^l =\{C_\alpha ^i|1\leq i\leq n_\alpha\}$. Each $C_\alpha ^i$ in the set is initialized as the root of a MST $\mathcal{T}_i^l$, and we initialize the weights in such a way that the weight $w_{ij}$ between the roots $C_\alpha ^i$ and the non-root vertices $C_j$ are the co-visibility counts ($\geq cov^\text{thr}$) and $0$ otherwise. Formally,
\begin{align}
\label{eq:mstweight}
w_{ij}=
\mathcal{X}_{\geq cov^\text{thr}}|\mathbf{P}_\alpha ^i \cap \mathbf{P}_j|, & \ 1\leq \forall j \neq i\leq m_l,
\end{align}
where $\mathcal{X}_{\geq cov^\text{thr}}$ denotes the corresponding indicator function. We follow~\cite{pettie2002randomized} to obtain the updated pose-graph $\mathcal{G}^{l'}=\{\mathcal{V}^{l^{'}}, \mathcal{E}^{l^{'}}\}$ where $\mathcal{V}^{l^{'}}\subseteq \mathcal{V}^l$. The camera poses are then initialized with those of the corresponding roots. 

Note that $\mathcal{T}_i^l$'s can possibly contain only a subset of the vertices $\mathcal{V}^l$ since there might be intra-block cameras which share no common viewpoints with those in $\mathcal{C}_\alpha ^l$. Poses of these cameras are therefore initialized with the reference frame of the block. We define the first frame of each block (as is also the last frame of the previous block) as the {\it local reference frame} of $\mathcal{B}^l$, \ie, $I_r^l\triangleq I_1^l$. 

So far we have obtained the MSF $\mathcal{M}^l$ from $\mathcal{G}^l$ with disjoint $\mathcal{T}_i^l$'s properly defined, formally, $\mathcal{M}^l=\{\mathcal{T}_i^l|i=1,\cdots, n_\alpha \}\cup \mathcal{T}_r^l$. Based on the observation that camera motions are likely to be bounded within a small range given the local co-visibility condition defined in \S\ref{sec:partition}, we argue that our initialization mechanism feeds the local BA with initial values sufficiently close to the optimal solution. We can then adopt our proposed {\it self-adaptive Levenberg-Marquardt solver} (\S\ref{sec:lm}) on solving Eq.~\ref{eq:locrepro} for the set of camera poses relative to $C_r^l$. We compare our proposed approach with conventional local optimization with fixed-size partitions in Fig.~\ref{fig:localcomp}, it can be observed that our method requires significantly fewer iterations while producing comparable or better accuracy. The intra-block relative poses are fixed upon the termination of local BA and the block is globally aligned with peer blocks in the consequent stage. 
\subsection{Global Alignment}
\label{sec:global}

As the results of each local BA are provided, we are now ready to align the blocks progressively to ensure the global consistency. Blocks are densely connected by multiple camera poses by virtue of our partitioning scheme, allowing us to uncover more accurate inter-block transformations compared with single-keyframe-based conventional global BA. Moreover, in order to tackle the computational bottleneck of global optimization based on point-camera correspondences, we leverage single motion averaging on the camera poses shared in different blocks.



Once the local optimization is done for a block, all the previous blocks sharing common camera poses with it are globally aligned. Specifically, several estimations of the same camera pose are derived from different blocks such that we can exploit Eq.~\ref{eq:rotavg} to measure the inter-block relative motions. Consider that two blocks $\mathcal{B}^l = \{\mathcal{C}^l, \mathcal{P}^l\}$ and $\mathcal{B}^{l'}=\{\mathcal{C}^{l'}, \mathcal{P}^{l'}\}$ share a common set ${}_l^{l'} \mathcal{C}$ of camera poses, where ${}_l^{l'} \mathcal{C}=\mathcal{C}^l \cap \mathcal{C}^{l'}$. Denote $\widetilde{C}_i^l$ and $\widetilde{C}_i^{l'}$ for the camera poses relative to the corresponding reference frames $C_r^l$ and $C_r^{l'}$, we thus have
\begin{subequations}
\begin{align}
\small
    \label{eq:estparam}
    {{}_l^{l^{'}} \widetilde{\mathcal{C}}}&=\{\widetilde{C}_i^l, \widetilde{C}_i^{l'}|l\neq l^{'}; i = 1,\cdots,|{}_l^{l^{'}} \mathcal{C}|\},\\
        {{}_l^{l^{'}} \widetilde{\mathcal{R}}}&=\{\widetilde{R}_i^l, \widetilde{R}_i^{l'}|l\neq l^{'}; i = 1,\cdots,|{}_l^{l^{'}} \mathcal{C}|\},
        \end{align}
\end{subequations}
where ${{}_l^{l^{'}} \widetilde{\mathcal{R}}}$ denotes the set of measurements of relative rotations.
We further define the inter-block relative rotation, as we aim to solve in the global alignment, $ R_{ll'}=R_{l'} R_l^\top$ from $\mathcal{B}^l$ to $\mathcal{B}^{l'}$, where $R_{l}$ and $R_{l'}$ are the \textit{pseudo} absolute rotations of the corresponding reference frames. Note that we never require the absolute rotation parameters in our algorithm. It thus follows that the estimated relative rotation $\widetilde{R}_{ll'}$ can be represented as 
\begin{equation}
    \widetilde{R}_{ll'}^i= \Big(\tilde{R_i} \tilde{R_{l'}}^{\top} \Big)^\top \Big(\tilde{R_i} \tilde{R_l}^\top\Big)=\widetilde{R}_i^{{l'}^\top} \widetilde{R}_i^l.
\end{equation}

Then we can solve the inter-block relative rotation $ R_{ll'}$ by rewriting Eq.~\ref{eq:rotavg} into the following form, 
\begin{equation}
\small
\label{eq:twoobj}
\argmin _{R_{ll'}\in \mathbb{SO}(3)}\sum_{i=1}^{|{}_l^{l^{'}} \mathcal{C}|}d\Big(\widetilde{R}_{ll'}^i , R_{ll'}\Big)^2.
\end{equation}

We are now ready to generalize Eq.~\ref{eq:twoobj} into the global optimization with all of the blocks where the local optimizations have been finished with the updated relative motion parameters. To find the set of aligned rotations $\mathcal{R}_{m_b}$ for $m_b$ blocks, where $\mathcal{R}_{m_b}=\{R_{jj'}|1\leq j\neq j' \leq m_b\}$, we minimize the global rotation estimation error by solving
\begin{equation}
\vspace{-2pt}
\label{eq:globalraobj}
    \argmin_{R_{jj'}\in \mathbb{SO}(3)}\sum_{j\neq j'}^{m_b} \sum_{i=1}^{|{}_j^{j'} \mathcal{C}|} d\Big(\widetilde{R}_{jj'}^i,R_{jj'}\Big)^2.
\end{equation}

 We argue that $d_{\angle}\big(\widetilde{R}_{jj'}^i,R_{jj'}\big) \in [0,\pi ], \forall i,j,j'$ to guarantee the global convergence and follow the algorithm in \cite{manton2004globally} for the optimization. In practice, the global alignment runs in parallel with the local optimization once a block is ready and Eq.~\ref{eq:globalraobj} is iteratively optimized only with the latest rotation parameters of the common cameras. 
 
  \begin{figure}[t!]
\begin{center}
\subfloat [Total Re-projection Error]{
\includegraphics[width=.48\linewidth]{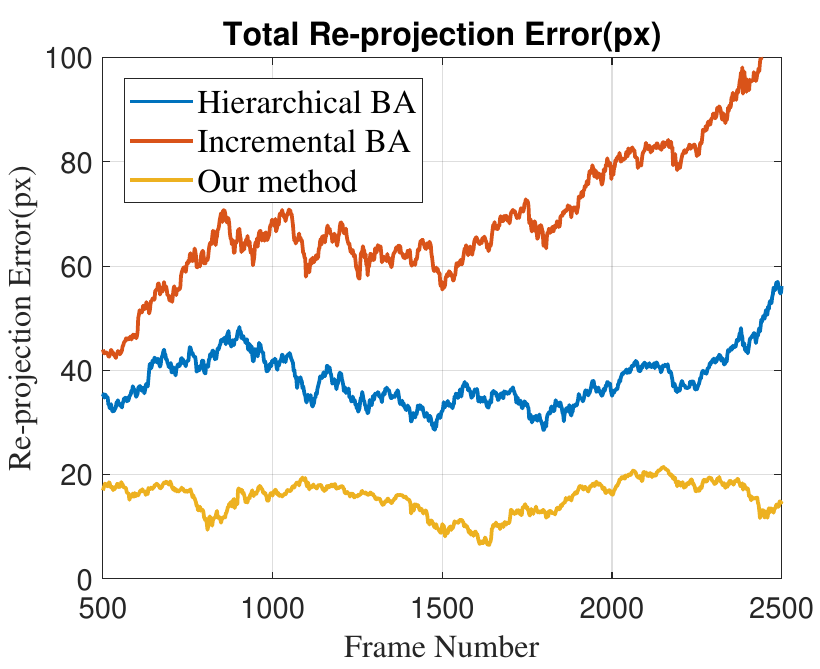}
\label{fig:totalerr}
}
\subfloat [Global Alignment Time]{
\includegraphics[width=.49\linewidth]{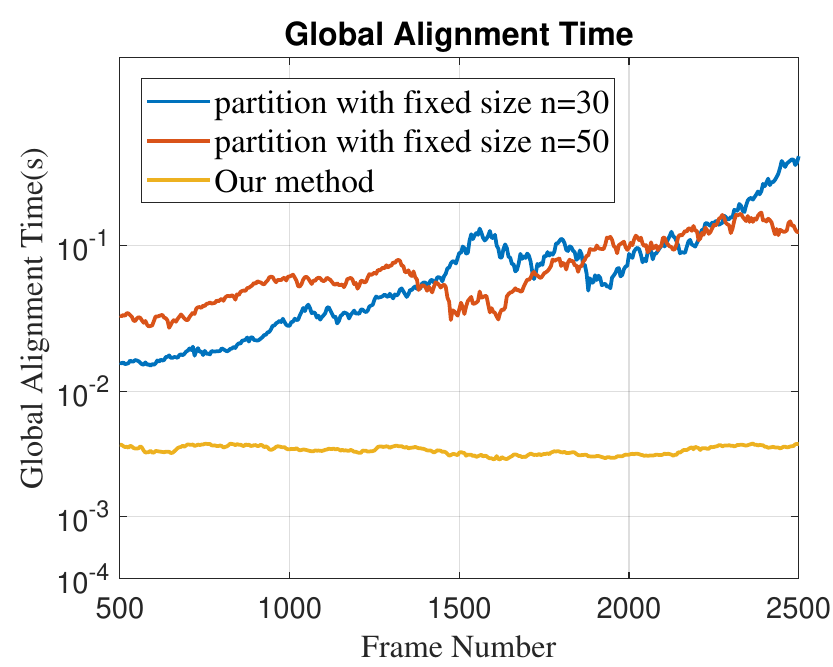}
\label{fig:gbatime}
}
\caption{\small Comparison with conventional global BA on the accuracy and runtime. (a) shows the accumulated re-projection comparison with conventional hierarchical BA and incremental BA methods. (b) shows the run-time comparison with conventional keyframe-based global BA with fixed-size partitions.The experiments are conducted on Seq.02 of KITTI Odometry~\cite{Geiger2013c}. Results on the whole sequence is available in the supplementary materials.}
\end{center}
\vspace{-0.7cm}
\end{figure}

 Compared with conventional global BA based on point-camera correspondences, our algorithm can achieve a much lower computational complexity and almost constant run-time with increasing number of image frames. For on-line systems where input image frames are sequentially processed, the inter-block global co-visibility stays on a stable level. Fig.~\ref{fig:totalerr} shows that the total re-projection error increases in a much lower rate compared to incremental and keyframe-based hierarchical BA. In Fig.~\ref{fig:gbatime} it can be observed that our global alignment is constantly 60-150x faster than that of conventional global BA based on point-camera correspondences. 
 
 According to the detection of the camera-block overlapping views based on global co-visibility, our algorithm can skip explicit loop closings to further accelerate the global alignment. In keyframe-based approaches, \XL{co-visible points are fused such that the robot observes a loop by recognizing a previously-visited location when sufficient inliers are detected}, followed by expensive global optimization iterations on all keyframes. By comparison, in our global alignment component the cameras showing in multiple blocks are optimized each time a new block is generated. Since the co-visibility is detected and stored in the partitioning stage, \XL{the global alignment directly conducts on the stored data association, avoiding the expensive on-line feature matching}. Furthermore, instead of processing the optimization on the whole loop, the blocks are aligned by averaging the individual intra-block `shared' camera poses. For example, if a new block observes similar scenes with one previous block, many camera poses in these two blocks will be shared and the update brought by the new block is automatically broadcast to all corresponding blocks by each individual common camera separately. \XL{Without the individual camera pose averaging and broadcasting scheme, our global alignment downgrades to conventional keyframed loop closure, only with denser data association, \ie, blocks observing similar scenes trigger the loop closure, followed by globally aligning all the reference frames of temporal blocks between the two corresponding reference frames. We demonstrate the performance of our global alignment scheme in Tables~\ref{table:tum} and~\ref{table:kittiquant}.}

\section{Self-Adaptive Numerical Solver}
\label{sec:anal}
\label{sec:lm}
To efficiently solve the local optimization problem defined in \S\ref{sec:local} with the intermediate results propagated from other blocks as the initial values, we propose a modified Levenberg-Marquardt solver assuming an initialization sufficiently close to the optimal solution set. Precisely, the step size of the iteration is updated according to the size of the block, achieving a quadratic convergence rate.

Letting $f(x) \equiv [f_1 (x), \cdots , f_m (x)]$ denote the stacked vector of re-projection errors where $f_i (x)$ represent the re-projection error on image frame $I_i$, to solve Eq.~\ref{eq:locrepro} we consider the equivalent non-linear system
\begin{equation}
\label{eq:zerofun}
    f(x)=0,\text{ s.t. }x\in \Omega,
\end{equation}
where $f$ is differentiable, and $\Omega \subseteq \mathcal{H} \subseteq \mathbb{R}^{6 m}\otimes \mathbb{R}^{3n}$ denotes the assumingly non-emepty solution set for a Hilbert space $\mathcal{H}$. We further define the related optimization problem
\begin{equation}
\small
\label{eq:obj}
    \min_{x\in \mathcal{H}} \phi(x), \text{ with }\phi:\mathcal{H} \mapsto \mathbb{R} \text{ given by }\phi(x)=\|f(x)\|_2^2.
\end{equation}

Let $\nabla f(x_{\boldsymbol{\cdot}})=J^\top (x_{\boldsymbol{\cdot}})$ denote the transpose of Jacobian of $f$ at $x_{\boldsymbol{\cdot}}$, then L-M is conventionally implemented to solve Eq.~\ref{eq:obj} by iteratively generating a sequence of convex sub-problems:
\begin{subequations}
\small
\begin{align}
    \min_{d\in \mathcal{H}} &\psi_k(d),\text{ with }\psi_k :\mathcal{H} \mapsto \mathbb{R} \text{ given by }\\
   \label{eq:dk} \psi_k(d)&=\|\nabla f(x_k)^\top d + f(x_k)\|_2^2 + \mu_k \|d\|_2^2,\\
    \label{eq:4c}(\nabla f(x_k) &\nabla f(x_k)^\top +\mu_k I) d_k=-\nabla f(x_k) f(x_k),
\end{align}
\end{subequations}
where $d$ in Eq.~\ref{eq:dk} is solved to compute the direction $d_k$ in Eq.~\ref{eq:4c}, $I\in \mathcal{H}\otimes \mathcal{H}$ denotes the identity matrix. $\mu_k\in \mathbb{R}$, namely Levenberg-Marquardt parameter, is updated with adaption to properly handle the cases where $\nabla f(x_k) \nabla f(x_k)^\top$ is singular. The Levenberg-Marquardt method is well-proven to achieve a quadratic convergence rate to a solution $x^*$ of Eq.~\ref{eq:zerofun} if $\nabla f(x^*)$ is nonsingular, where the nonsingularity also implies that the solution to Eq.~\ref{eq:obj} must be locally unique. However, the assumption of nonsingularity fails for the conventional BA problems.  

Motivated by~\cite{yamashita2001rate}, we assume a much weaker assumption, namely, {\it local error bound condition} than the nonsingularity of $\nabla f(x^*)$. Denote $\text{dist}(y, \Omega)=\inf_{x\in \Omega}\|y-x\|_2$, recall the local error bound condition as
\begin{equation}\small \label{eq:lcb}
    \exists \delta,\beta>0\text{ s.t. }\beta\ \text{dist}(x, \Omega) \leq \|f(x)\|_2, \hspace{5pt} \forall x\in B_\delta (x^*)\cap \Omega,
\end{equation}
where $B_\delta(x^*)=\{x\in\mathcal{H}|\text{dist}(x,x^*)\leq \delta\}$.


Given that the Eq.~\ref{eq:lcb} holds, the choice of Levenberg-Marquardt parameter $\mu_k$ plays a key role in the Levenberg-Marquardt convergence. We argue that the initialization of the iteration sequence defined in Eq.~\ref{eq:locrepro} is sufficiently close to the solution set $\Omega$ by propagating and broadcasting the common camera poses as illustrated in \S\ref{sec:local}. Letting $\mu_k = \alpha_k \|f(x_k)\|_2^{\lambda}$ in Eq.~\ref{eq:dk} thus leads to the quadratic rates of local convergence for Eq.~\ref{eq:obj}, where $\alpha_k$ is iteratively updated to control the \textit{trust region} radius of searching a satisfactory $d_k$. In details,
\begin{subequations}
\label{eq:search}
\small
\begin{align}
    \mu_k& = \alpha_k\|f(x_k)\|_2^\lambda,\\
    \Phi_k&=\phi(x_k)-\phi(x_k + d_k),\\
    \Psi_k&=\phi(x_k)-\min_{d_k \in \mathcal{H}} \|\nabla f(x_k)^\top d_k +f(x_k)\|_2^2,
\end{align}
\end{subequations}
where $\Phi_k$ and $\Psi_k$ can be considered as the actual and predicted reduction for the merit function, respectively. The parameter $\nu_k \in (0,1)$ is introduced to the iterative system in order to serve as an indicator of whether the current $d_k$ benefits towards the solution, further guiding the update rule of $\alpha_k$. Formally,
\begin{subequations}
\label{eq:nu}
\small
\begin{align}
    \nu_k  &=\dfrac{\Phi_k}{\Psi_k},\\
    \alpha_{k+1} &= \rho(\eta, \alpha_k, \nu_k;\nu),
\end{align}
\end{subequations}
where $\nu\in(0,1)$ is a given constant, $\eta >1$ is a positive integer, $\rho(\eta, \alpha_{\boldsymbol\cdot}, \nu_{\boldsymbol \cdot};\nu):\mathbb{N}_{>0} \times \mathbb{R}\times (0,1)\mapsto \mathbb{R}_{\geq 0}$ is continuous and nonnegative (here we let $\rho(\eta,\alpha_{\boldsymbol\cdot}, \nu_{\boldsymbol \cdot};\nu)$ be strictly positive by enforcing a sufficiently small lower bound $\xi>0$), defined as
\begin{equation}
\label{eq:rho}
    \rho(\eta,\alpha_k, \nu_k;\nu)=\alpha_k \max\{\xi,1-\eta^{\ceil{\frac{1}{\nu}}}(\nu_k-\nu)^{2\ceil{\log_2 \eta}-1}\}.
\end{equation}

Our formulation of Eq.~\ref{eq:rho} can be seen as a variant of the update rule in \cite{Fan2006} and \cite{Hei2003}, 
the proposed iteration scheme has the following desirable properties: (i) $\nu_k=\nu$ is the only inflection point, \ie$\frac{\partial^2 \rho_\nu}{\partial \nu_k^2}<0$ if $\nu_k < \nu$ and the opposite holds otherwise; (ii) $\rho_\nu(\eta, \alpha_k, \nu_k)>\alpha_k$ $\forall \eta,\alpha_k$, if $\nu_k <\nu$ and the opposite holds otherwise. Specifically, the properties imply that as $\nu_k < \nu$ and gets smaller (\eg close to $0$), $\alpha_k$ (and thus $\mu_k$) are enlarged at a higher rate to improve the searching for a better $d_k$; for the contrary case when $\nu_k$ is close to $1$, $\alpha_k$ can be legitimately reduced since $d_k$ is satisfactory. 
In addition, we let $\eta =m_l$ for the problem defined in Eq.~\ref{eq:locrepro} with $m_l$ image frames. Since a high value of $n$ foresees a slow convergence, we thus add the power term $(2\ceil{\log_2 \eta}-1)$ such that the sensitivity of $\rho_\nu$ with regards to $\nu_k$ increases with adaption for a greater $n$ to preserve the convergence speed for large-scale problems. The convergence properties are shown similar to~\cite{Fan2005,Fan2006, Kanzow2005b} in the supplementary materials.

The local pose refinement is conducted in an incremental BA manner with our proposed Levenberg-Marquardt solver. Our algorithm improves the local optimization in two-fold: First, with our local co-visibility threshold defined for the partitioning scheme, it is guaranteed that a majority of the 3D points are observable by a sufficient amount of cameras to provide the high accuracy. Second, the inter-block propagation of the intermediate computational values greatly benefits the convergence of local Levenberg-Marquardt iterations, by providing the close-to-optimal initialization. 


\section{Experimental Results}
\label{sec:experiment}
\subsection{Implementation Details}
\label{sec:implem}

\textit{System Configuration} All of our experiment results are achieved from a PC with Intel(R) i7-7700 3.6GHz processors, 8 threads and 64GB memory with GTX1060 6GB GPU for SIFT~\cite{Lowe2004c} extraction and matching. 


\textit{Methods and Datasets} We compare our proposed method with several state-of-the-art SLAM systems including: DVO-SLAM~\cite{Kerl2013a}, Kintinuous~\cite{Whelan2012a}, Elastic Fusion~\cite{Whelan2015b},LDSO~\cite{gao2018ldso} and ORB-SLAM~\cite{Mur-Artal2015a}. We also assess keyframe-based offline SfM systems including ENFT-SFM~\cite{Zhang2016a} and VisualSFM~\cite{wu2011visualsfm} in our comparison for a comprehensive evaluation. The methods are tested on TUM-RGBD~\cite{Handa2014b} and KITTI Odometry~\cite{Geiger2013c} datasets. For all the sequences on both datasets we use SIFT~\cite{Lowe2004c} and ORB~\cite{rublee2011orb} features for feature matching followed by RANSAC~\cite{fischler1981random} iterations to remove feature correspondence outliers, where we only use GPU for SIFT feature correspondence computation in parallel. We have implemented and modified Ceres Solver (\cite{ceres-solver, kummerle2011g}) for local BA, and followed the algorithm in~\cite{manton2004globally} for the global alignment.

\begin{figure*}[!ht]
\begin{center}
\subfloat	[Sequence 02]	{
\includegraphics[width=.342\linewidth]{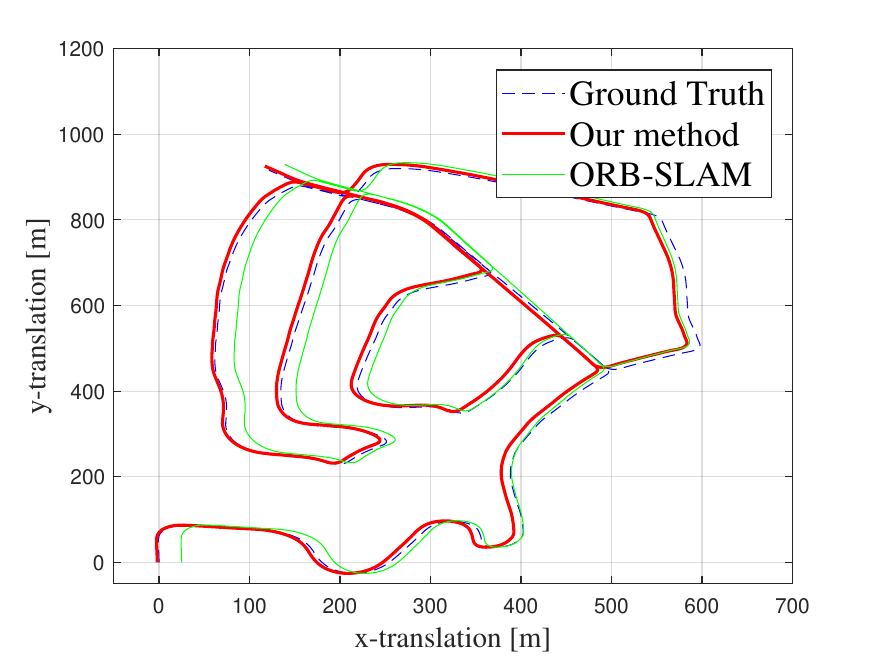}
}
\subfloat  [Sequence 08]    {
\includegraphics[width=.31\linewidth]{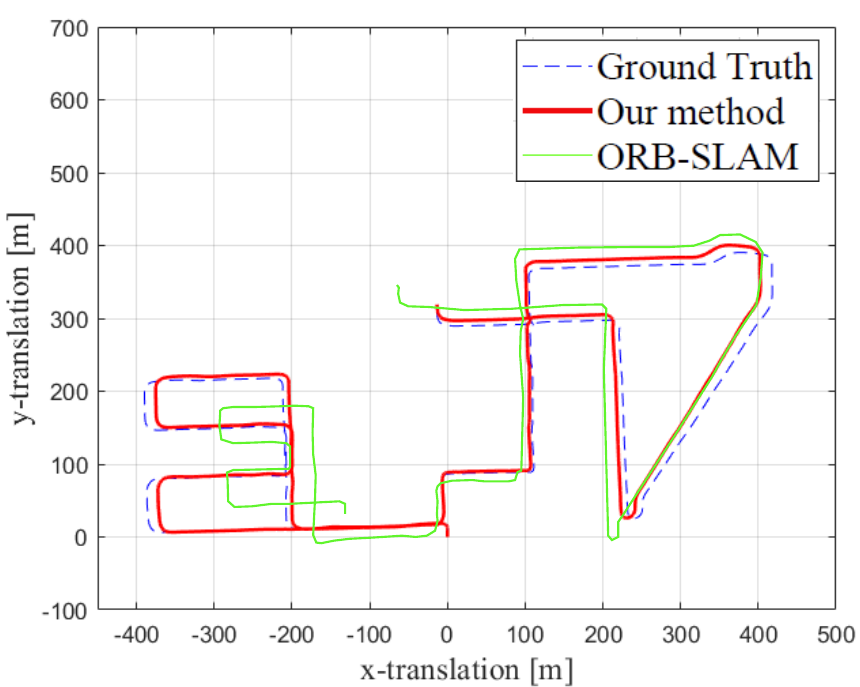}
}
\subfloat   [Sequence 10]   {
\includegraphics[width=.31\linewidth]{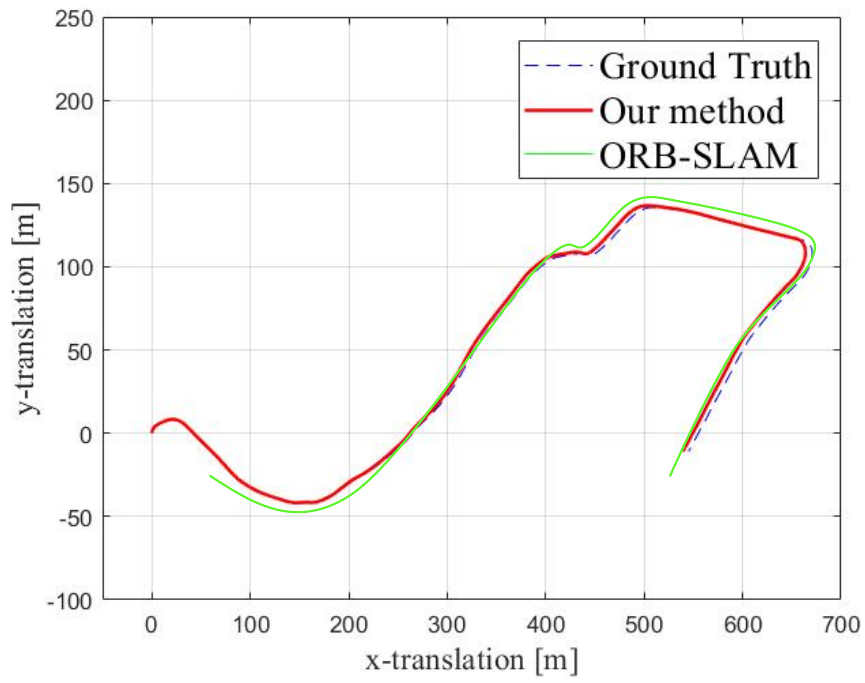}
}
\caption[Comparison with ORB-SLAM on Seq.02, 08, 10 from KITTI Odometry.]{\small Our proposed approach shows a trajectory estimation of comparably high quality in large scale outdoor scenes. Note on Seq.02 it is clearly shown that ORB-SLAM (in green) suffers from accumulated drifts although the trajectory contains many loops. Results on the rest of the dataset are provided in the supplementary materials where our approach outperforms ORB-SLAM on all of the sequences.}
\label{fig:3seq}
\end{center}
\vspace{-12pt}
\end{figure*}

\subsection{Evaluations on Real-World Data}
\label{sec:result}
\begin{table}[!t]
\setlength{\belowcaptionskip}{-25pt}
\begin{center}
\resizebox{\linewidth}{!}{%
\begin{tabular}{r | @{\hspace{0.75mm}}c@{\hspace{.75mm}} @{\hspace{0.75mm}}c@{\hspace{.75mm}} @{\hspace{0.75mm}}c@{\hspace{.75mm}} @{\hspace{0.75mm}}c@{\hspace{.75mm}} @{\hspace{0.75mm}}c@{\hspace{.75mm}} @{\hspace{0.75mm}}c@{\hspace{.75mm}} |@{\hspace{0.75mm}}c@{\hspace{.75mm}} @{\hspace{0.75mm}}c@{\hspace{.75mm}} @{\hspace{0.75mm}}c@{\hspace{.75mm}}}

\hline
& \thead{DVO\\SLAM \\~\cite{Kerl2013a}} & \thead{Kinti-\\nuous\\~\cite{Whelan2012a}} & \thead{Elastic\\ Fusion\\~\cite{Whelan2015b}} & \thead{ORB\\SLAM\\~\cite{Mur-Artal2015a}}& \thead{ENFT\\SFM\\~\cite{Zhang2016a}} &\thead{Visual\\SFM\\~\cite{wu2011visualsfm}}&\thead{OURS\\(SIFT)} &\thead{OURS\\(LC)} &\thead{OURS\\(ORB)}\\
\hline \hline 
\footnotesize{fr1\_desk} & 2.1 &3.7&2.0&\textcolor{cyan}{\textbf{1.7}}&2.7&2.7&\textcolor{red}{\textbf{1.6}}&\textcolor{red}{\textbf{1.6}} &\textcolor{cyan}{\textbf{1.7}}\\
\hline
\small{fr1\_desk2} & 4.6 & 7.1&4.8&{2.9}*&3.6*&-&\textcolor{red}{\textbf{1.5}}&\textcolor{cyan}{\textbf2.0} & 2.3\\
\hline
\small{fr1\_room} & 4.3 & 7.5&6.8&\textcolor{cyan}{\textbf{3.4}}*&5.3*&-&\textcolor{red}{\textbf{2.9}}& \textcolor{cyan}{\textbf{3.4}} & 3.2\\
\hline
\small{fr2\_desk} & 1.7 & 3.4 & 7.1&\textcolor{red}{\textbf{0.8}}&2.3&1.7&1.5&\textcolor{cyan}{\textbf{1.2}} & 2.0\\
\hline
\small{fr2\_xyz} & 1.8&2.9&1.1&\textcolor{red}{\textbf{0.3}}&0.8&\textcolor{cyan}{\textbf{0.7}}&\textcolor{red}{\textbf{0.3}}&\textcolor{red}{\textbf{0.3}} & \textcolor{red}{\textbf{0.3}}\\
\hline
\small{fr3\_office} & 3.5 & 3.0 & 1.7&3.5&1.2&\textcolor{cyan}{\textbf{1.1}}&\textcolor{red}{\textbf{0.8}}&1.3 & \textcolor{cyan}{\textbf{1.1}}\\
\hline
\small{fr3\_nst\_near} & 1.8 & 3.1&1.6&\textcolor{cyan}{\textbf{1.4}}&1.9&\textcolor{red}{\textbf{1.1}}&\textcolor{red}{\textbf{1.1}}&1.6 &\textcolor{cyan}{\textbf{1.4}}\\
\hline
\end{tabular}}
\caption{\small Trajectory estimation RMSE (cm) on the TUM RGB-D \cite{Handa2014b} dataset. Results with (*) are experimental results based on 5-execution medians from running open-source codes provided by the authors. OURS(LC) reports the results of our approach using general pose-graph optimization framework with loop closure scheme. The first three systems additionally use depth data. OURS(ORB) reports the results with ORB feature.}
\label{table:tum}
\end{center}

\end{table}
\textbf{Evaluation on TUM RGB-D.}
As a challenging dataset for monocular online systems, the TUM RGB-D dataset contains specific RGB-D sequences including rotation-only motion and textureless surface scanning in a small scene, captured by Kinect. As we do not rely on depth information in our system, we run experiments on a selected part of the dataset which are suitable for monocular systems and evaluate the results of camera trajectory against state-of-the-art systems. The quantitative comparison results can be found in Table~\ref{table:tum}. Note that DVO-SLAM, Kintinuous and Elastic Fusion use the depth data in their experiments, ENFT-SFM and VisualSFM are mostly processed offline. We quote the experiment results of VisualSFM from~\cite{Zhang2016a}, where the authors state that the results are achieved from their manually-extracted keyframes off-line. Though our approach achieves clear outperformances on most of the sequence, ORB-SLAM shows strong robustness dealing with dynamic objects (in \textit{fr2\_desk} sequence).

\textbf{Evaluation on KITTI Odometry.} The KITTI dataset contains 11 video sequences captured by a fast-moving vehicle over long camera trajectories. In the experiments it has been shown that systems like ORB-SLAM which relies heavily on explicit loop closures suffer on several sequences. \XL{Our proposed approach ensures global consistency by consecutively aligning the blocks and} convincingly shows the robustness on large-scale outdoor scenes.



\begin{table}[h!]
\setlength{\belowcaptionskip}{-20pt}
\begin{center}
\resizebox{\linewidth}{!}{
\begin{tabular}{@{\hspace{0.75mm}}r@{\hspace{0.95mm}}|@{\hspace{0.75mm}}c@{\hspace{.75mm}} @{\hspace{0.75mm}}c@{\hspace{.75mm}} @{\hspace{0.275mm}}c@{\hspace{.275mm}} @{\hspace{0.75mm}}c@{\hspace{.75mm}} | @{\hspace{0.75mm}}c@{\hspace{0.75mm}} @{\hspace{0.75mm}}c@{\hspace{0.75mm}} @{\hspace{0.75mm}}c@{\hspace{0.75mm}} @{\hspace{0.75mm}}c@{\hspace{0.275mm}}}
\hline
 & {\thead{ENFT\\ SFM\\ \cite{Zhang2016a}}} & \thead{Visual\\ SFM\\ \cite{wu2011visualsfm}} &\thead{LDSO\\ \cite{gao2018ldso}} &\thead{ORB\\ SLAM\\ \cite{Mur-Artal2015a}} & \thead{OURS\\(SIFT)} & \thead{OURS\\(LC)} & \thead{OURS\\(ORB)} & \thead{OURS \\ (DL)} \\
\hline \hline
00&4.8&2.8 (3.7\%)&9.3&5.3 /15.26s&\textcolor{red}{\textbf{2.5}} /5.65s & 4.2 /12.7s & 4.9 /3.42s & 2.8 /8.39s\\
\hline
01&57.2 &52.3 (12.5\%)& \textcolor{red}{\textbf{11.7}} & - & 37.4 /14.96s & 48.2 /10.24s &  45.9 /6.77s & 39.3 /17.2s \\
\hline
02&28.3&1.8 (4.5\%)& 32.0&21.3 /21.4s&\textcolor{red}{\textbf{15.1}} /7.26s & 24.9 /25.6s & 18.2 /4.43s & 15.9 /13.2s \\
\hline
03&2.9&0.3 (12.0\%)&2.9 &1.5 /1.97s&\textcolor{red}{\textbf{1.2}} /0.93s & 1.6 /0.90s & 1.5 /0.78s & 1.2 /1.44s\\
\hline
04&0.7&0.8 (23.4\%)&1.2&1.6 /1.02s&\textcolor{red}{\textbf{0.3}} /0.17s & 1.0 /1.9s & 0.9 /0.14s & 0.3 /0.59s\\
\hline
05&3.5&9.8 (7.4\%)&5.1&2.9 /6.93s&\textcolor{red}{\textbf{2.8}} /2.35s & 3.4 /6.50s & 2.9 /4.87s & 3.2 /6.23s\\
\hline
06&14.4&8.6 (7.4\%)&13.6&12.3 /2.87s&\textcolor{red}{\textbf{6.3}} /0.84s & 10.7 /1.92s & 8.2 /0.65s & 9.2 /3.75s\\
\hline
07&\textcolor{red}{\textbf{2.0}}&3.9 (7.8\%)&3.0&2.3 /1.22s&\textcolor{red}{\textbf{2.0}} /0.91s & 2.5 /1.43s &2.2 /0.70s & 2.2 /1.54s\\
\hline
08&28.3&0.8 (0.9\%)&129.0&46.7 /20.9s&\textcolor{red}{\textbf{19.7}} /15.87s &25.7 /24.0s & 33.6 /10.09s & 21.9 /21.4s\\
\hline
09&\textcolor{red}{\textbf{5.9}}&0.9 (4.9\%)&21.7&6.6 /4.47s&6.2 /2.54s &6.2 /4.65s & 6.5 /1.96s & 6.2 /5.57s\\
\hline
10&18.5&5.7 (6.5\%)&17.4&8.8 /3.72s&\textcolor{red}{\textbf{5.2}} /1.02s & 6.1 /2.85s & 7.6 /0.69s & 6.0 /2.12s\\
\hline
\end{tabular}}
\caption{\small Trajectory estimation RMSE(m) and pose estimation runtime on KITTI Odometry. For VisualSFM the number in the parenthesis represents the map completeness ratio and it is 100\% for all other methods. For ORB-SLAM and our approach the second number represents the BA runtime. In addition, OURS(DL) replaces our modified LM solver with conventional Dogleg algorithm~\cite{ceres-solver}. Note that ENFT-SFM and VisualSFM are offline SfM systems and the latter requires preprocessings. ORB-SLAM fails to initialize on Seq.01 so we leave the result blank.}
\label{table:kittiquant}
\end{center}
\end{table}

\textit{Quantitative Comparison} For the quantitative comparisons shown in Table.~\ref{table:kittiquant}. We quote the results of VisualSFM from~\cite{Zhang2016a}, where the authors pointed out that the trajectories are overly long for VisualSFM, resulting in disconnected blocks of trajectories. Hence the RMSE results are calculated by taking the largest component of the trajectory. Moreover, ORB-SLAM fails to process Seq.01 due to lost of tracking at the beginning of the video. \XL{We have also observed the average amounts of matched features over all sequences are 1126 and 837 per frame for SIFT and ORB, with inlier rates 0.6245 and 0.4528, respectively. In our experiments, the total feature extraction and matching time are 0.016s and 0.006s per frame for SIFT and ORB, respectively, on GPU. Furthermore, as the amount of post-RANSAC ORB is only about half of that of SIFT, our approach with ORB yields a lower accuracy though the pose estimation is faster.}

\textit{Qualitative Comparison} Though ORB-SLAM is able to correctly detect and close the loop in Seq.02, it still shows large accumulated drift. Our approach shows a clearly more accurate estimation on Seq.08 and Seq.10 (in Fig.~\ref{fig:3seq}). Seq.08 does not contain loops so ORB-SLAM suffers from serious drifts both on translations and scale. While on Seq.10 ORB-SLAM has lost tracking from the beginning of the sequence and also shows accumulated drift by the end of trajectory estimation. The rest of the qualitative comparisons are given in the supplementary materials, where our approach outperforms ORB-SLAM on all of the sequences.



\textit{Modified LM Solver} We replace our modified LM solver with conventional  Dogleg algorithm~\cite{ceres-solver} in local optimization and achieve better efficiency as shown. 

\section{Conclusion}
\label{sec:conclusion}
\XL{In this paper, we propose a hybrid 
real-time camera pose estimation formulation with online partitioning. Our partitioning scheme ensures the strong connectivity between the cameras such that the pose estimation accuracy is significantly improved. With the propagation of intermediate computation values, our modified Levenberg-Marquardt solver further enhances the local BA efficiency. The inter-block dense data association enables us to introduce single rotation averaging for the global alignment, as to avoid the high computational costs and explicit loop closures in conventional global BA frameworks. Clear outperformances against state-of-the-art monocular systems with conventional BA further demonstrate the practicality and robustness of our proposed approach.
}

\vspace{0.3em}
\noindent
{
\bibliographystyle{IEEEtran}
\bibliography{20RAL}

\begin{thebibliography}{10}
\providecommand{\url}[1]{#1}
\csname url@samestyle\endcsname
\providecommand{\newblock}{\relax}
\providecommand{\bibinfo}[2]{#2}
\providecommand{\BIBentrySTDinterwordspacing}{\spaceskip=0pt\relax}
\providecommand{\BIBentryALTinterwordstretchfactor}{4}
\providecommand{\BIBentryALTinterwordspacing}{\spaceskip=\fontdimen2\font plus
\BIBentryALTinterwordstretchfactor\fontdimen3\font minus
  \fontdimen4\font\relax}
\providecommand{\BIBforeignlanguage}[2]{{%
\expandafter\ifx\csname l@#1\endcsname\relax
\typeout{** WARNING: IEEEtran.bst: No hyphenation pattern has been}%
\typeout{** loaded for the language `#1'. Using the pattern for}%
\typeout{** the default language instead.}%
\else
\language=\csname l@#1\endcsname
\fi
#2}}
\providecommand{\BIBdecl}{\relax}
\BIBdecl

\bibitem{Indelman2012a}
V.~Indelman, R.~Roberts, C.~Beall, and F.~Dellaert, ``{Incremental Light Bundle
  Adjustment},'' in \emph{BMVC}, 2012.

\bibitem{Vo2016b}
M.~Vo, S.~G. Narasimhan, and Y.~Sheikh, ``{Spatiotemporal Bundle Adjustment for
  Dynamic 3D Reconstruction},'' in \emph{CVPR}, 2016.

\bibitem{Sibley2009b}
D.~Sibley and C.~Mei, ``{Adaptive relative bundle adjustment},'' in \emph{RSS},
  2009.

\bibitem{Wu2013b}
C.~Wu, ``{Towards linear-time incremental structure from motion},'' in
  \emph{3DV}, 2013.

\bibitem{schonberger2016structure}
J.~L. Schonberger and J.-M. Frahm, ``Structure-from-motion revisited,'' in
  \emph{CVPR}, 2016.

\bibitem{Liu2018a}
H.~Liu, M.~Chen, G.~Zhang, H.~Bao, and Y.~Bao, ``{ICE-BA: Incremental,
  Consistent and Efficient Bundle Adjustment for Visual-Inertial SLAM},''
  \emph{CVPR}, 2018.

\bibitem{Konolige2008b}
K.~Konolige and M.~Agrawal, ``{FrameSLAM: From bundle adjustment to real-time
  visual mapping},'' \emph{TRO}, 2008.

\bibitem{kaess2008}
M.~Kaess, A.~Ranganathan, and F.~Dellaert, ``i{SAM}: Incremental smoothing and
  mapping,'' in \emph{TRO}, 2008.

\bibitem{ila2017}
V.~Ila, L.~Polok, M.~Solony, and P.~Svoboda, ``{SLAM}++-a highly efficient and
  temporally scalable incremental slam framework,'' \emph{IJRR}, 2017.

\bibitem{Shum1999}
H.~Shum, Q.~Ke, and Z.~Zhang, ``{Efficient bundle adjustment with virtual key
  frames: a hierarchical approach to multi-frame structure from motion},'' in
  \emph{CVPR}, 1999.

\bibitem{Maier2014d}
R.~Maier, J.~Sturm, and D.~Cremers, ``{Submap-based bundle adjustment for 3D
  reconstruction from RGB-D data},'' in \emph{BMVC}, 2014.

\bibitem{Ni2007a}
K.~Ni, D.~Steedly, and F.~Dellaert, ``{Out-of-core bundle adjustment for
  large-scale 3D reconstruction},'' in \emph{ICCV}, 2007.

\bibitem{steedly2003spectral}
D.~Steedly, I.~Essa, and F.~Dellaert, ``Spectral partitioning for structure
  from motion.''\hskip 1em plus 0.5em minus 0.4em\relax Georgia Institute of
  Technology, 2003.

\bibitem{havlena2010efficient}
M.~Havlena, A.~Torii, and T.~Pajdla, ``Efficient structure from motion by graph
  optimization,'' in \emph{ECCV}, 2010.

\bibitem{cui2017hsfm}
H.~Cui, X.~Gao, S.~Shen, and Z.~Hu, ``{HS}f{M}: Hybrid structure-from-motion,''
  in \emph{CVPR}, 2017.

\bibitem{grisetti2012robust}
G.~Grisetti, R.~K{\"u}mmerle, and K.~Ni, ``Robust optimization of factor graphs
  by using condensed measurements,'' in \emph{ICIRS}, 2012.

\bibitem{ni2007tectonic}
K.~Ni, D.~Steedly, and F.~Dellaert, ``Tectonic {SAM}: Exact, out-of-core,
  submap-based {SLAM},'' in \emph{ICRA}, 2007.

\bibitem{cui2015}
Z.~Cui and P.~Tan, ``Global structure-from-motion by similarity averaging,'' in
  \emph{ICCV}, 2015.

\bibitem{zhu2018}
S.~Zhu, R.~Zhang, L.~Zhou, T.~Shen, T.~Fang, P.~Tan, and L.~Quan, ``Very
  large-scale global {S}f{M} by distributed motion averaging,'' in \emph{CVPR},
  2018.

\bibitem{zhu2017parallel}
S.~Zhu, T.~Shen, L.~Zhou, R.~Zhang, J.~Wang, T.~Fang, and L.~Quan, ``Parallel
  structure from motion from local increment to global averaging,'' \emph{arXiv
  preprint arXiv:1702.08601}, 2017.

\bibitem{chatterjee2013efficient}
A.~Chatterjee and V.~Madhav~Govindu, ``Efficient and robust large-scale
  rotation averaging,'' in \emph{ICCV}, 2013.

\bibitem{chatterjee2018robust}
A.~Chatterjee and V.~M. Govindu, ``Robust relative rotation averaging,''
  \emph{PAMI}, 2018.

\bibitem{hartley2013rotation}
R.~Hartley, J.~Trumpf, Y.~Dai, and H.~Li, ``Rotation averaging,'' \emph{IJCV},
  2013.

\bibitem{fredriksson2012simultaneous}
J.~Fredriksson and C.~Olsson, ``Simultaneous multiple rotation averaging using
  lagrangian duality,'' in \emph{ACCV}, 2012.

\bibitem{cui2017csfm}
H.~Cui, S.~Shen, X.~Gao, and Z.~Hu, ``{CS}f{M}: community-based structure from
  motion,'' in \emph{ICIP}, 2017.

\bibitem{eriksson2016consensus}
A.~Eriksson, J.~Bastian, T.-J. Chin, and M.~Isaksson, ``A consensus-based
  framework for distributed bundle adjustment,'' in \emph{CVPR}, 2016.

\bibitem{zhang2017distributed}
R.~Zhang, S.~Zhu, T.~Fang, and L.~Quan, ``Distributed very large scale bundle
  adjustment by global camera consensus,'' in \emph{ICCV}, 2017.

\bibitem{locherprogsfm}
A.~Locher, M.~Havlena, and L.~V. Gool, ``Progressive structure from motion,''
  in \emph{ECCV}, 2018.

\bibitem{alvarowhyba}
A.~P. Bustos, T.-J. Chin, A.~Eriksson, and I.~Reid, ``Visual {SLAM}: Why bundle
  adjustment,'' in \emph{ICRA}, 2019.

\bibitem{grove1973conjugatec}
K.~Grove and H.~Karcher, ``How to conjugatec 1-close group actions,''
  \emph{Mathematische Zeitschrift}, vol. 132, no.~1, pp. 11--20, 1973.

\bibitem{Levenberg1944}
K.~Levenberg, ``{A method for the solution of certain non-linear problems in
  least squares},'' \emph{Quarterly of Applied Mathematics}, 1944.

\bibitem{Geiger2013c}
A.~Geiger, P.~Lenz, C.~Stiller, and R.~Urtasun, ``{Vision meets robotics: The
  KITTI dataset},'' \emph{IJRR}, 2013.

\bibitem{kruskal1956shortest}
J.~B. Kruskal, ``On the shortest spanning subtree of a graph and the traveling
  salesman problem,'' \emph{Proceedings of the AMS}, vol.~7, no.~1, 1956.

\bibitem{pettie2002randomized}
S.~Pettie and V.~Ramachandran, ``A randomized time-work optimal parallel
  algorithm for finding a minimum spanning forest,'' \emph{SIAM Journal on
  Computing}, vol.~31, no.~6, 2002.

\bibitem{manton2004globally}
J.~H. Manton, ``A globally convergent numerical algorithm for computing the
  centre of mass on compact lie groups,'' in \emph{ICARCV}, 2004.

\bibitem{yamashita2001rate}
N.~Yamashita and M.~Fukushima, ``On the rate of convergence of the
  {Levenberg-Marquardt} method,'' in \emph{Topics in Numerical Analysis}.\hskip
  1em plus 0.5em minus 0.4em\relax Springer, 2001.

\bibitem{Fan2006}
J.~Fan and J.~Pan, ``Convergence properties of a self-adaptive
  {Levenberg-Marquardt} algorithm under local error bound condition,''
  \emph{Computational Optimization and Applications}, vol.~34, no.~1, 2006.

\bibitem{Hei2003}
L.~Hei, ``A self-adaptive trust region algorithm,'' \emph{Journal of
  Computational Mathematics}, 2003.

\bibitem{Fan2005}
J.~Fan and Y.~Yuan, ``On the quadratic convergence of the {Levenberg-Marquardt}
  method without nonsingularity assumption,'' \emph{Computing}, vol.~74, no.~1,
  2005.

\bibitem{Kanzow2005b}
C.~Kanzow, N.~Yamashita, and M.~Fukushima, ``{Levenberg-Marquardt methods with
  strong local convergence properties for solving nonlinear equations with
  convex constraints},'' \emph{Journal of Computational and Applied
  Mathematics}, 2005.

\bibitem{Lowe2004c}
D.~G. Lowe, ``{Distinctive image features from scale-invariant keypoints},''
  \emph{IJCV}, 2004.

\bibitem{Kerl2013a}
C.~Kerl, J.~Sturm, and D.~Cremers, ``{Dense visual SLAM for RGB-D cameras},''
  in \emph{ICIRS}, 2013.

\bibitem{Whelan2012a}
T.~Whelan, M.~Kaess, M.~Fallon, H.~Johannsson, J.~Leonard, and J.~McDonald,
  ``{Kintinuous: Spatially Extended KinectFusion},'' \emph{Computer Science and
  Artificial Intelligence Laboratory Technical Report}, 2012.

\bibitem{Whelan2015b}
T.~Whelan, S.~Leutenegger, R.~{Salas Moreno}, B.~Glocker, and A.~Davison,
  ``{ElasticFusion: Dense SLAM Without A Pose Graph},'' in \emph{RSS}, 2015.

\bibitem{gao2018ldso}
X.~Gao, R.~Wang, N.~Demmel, and D.~Cremers, ``{LDSO}: Direct sparse odometry
  with loop closure,'' in \emph{IROS}, 2018.

\bibitem{Mur-Artal2015a}
R.~Mur-Artal, J.~M. Montiel, and J.~D. Tardos, ``{ORB-SLAM: A Versatile and
  Accurate Monocular SLAM System},'' \emph{TRO}, 2015.

\bibitem{Zhang2016a}
G.~Zhang, H.~Liu, Z.~Dong, J.~Jia, T.~T. Wong, and H.~Bao, ``{Efficient
  Non-Consecutive Feature Tracking for Robust Structure-From-Motion},''
  \emph{TIP}, 2016.

\bibitem{wu2011visualsfm}
C.~Wu, ``{VisualSFM}: A visual structure from motion system,'' 2011.

\bibitem{Handa2014b}
A.~Handa, T.~Whelan, J.~McDonald, and A.~J. Davison, ``{A benchmark for RGB-D
  visual odometry, 3D reconstruction and SLAM},'' in \emph{ICRA}, 2014.

\bibitem{rublee2011orb}
E.~Rublee, V.~Rabaud, K.~Konolige, and G.~R. Bradski, ``{ORB}: An efficient
  alternative to {SIFT or SURF},'' in \emph{ICCV}, 2011.

\bibitem{fischler1981random}
M.~A. Fischler and R.~C. Bolles, ``Random sample consensus: a paradigm for
  model fitting with applications to image analysis and automated
  cartography,'' \emph{Communications of the ACM}, vol.~24, no.~6, 1981.

\bibitem{ceres-solver}
S.~Agarwal and K.~M. et~al., ``Ceres solver,'' \url{http://ceres-solver.org}.

\bibitem{kummerle2011g}
R.~K{\"u}mmerle, G.~Grisetti, H.~Strasdat, K.~Konolige, and W.~Burgard, ``g 2
  o: A general framework for graph optimization,'' in \emph{ICRA}, 2011.

\end{thebibliography}
}
\end{document}